\documentclass{article}
\usepackage{multirow}

 \usepackage[main, final]{neurips_2026}

\makeatletter
\def\@noticestring{}
\makeatother

\usepackage[utf8]{inputenc}
\usepackage[T1]{fontenc}
\usepackage{hyperref}
\usepackage{url}
\usepackage{booktabs}
\usepackage{amsfonts}
\usepackage{nicefrac}
\usepackage{microtype}
\usepackage{xcolor}
\usepackage{enumitem}
\usepackage{graphicx}
\usepackage{float}
\usepackage{tabularx}
\usepackage[T1]{fontenc}
\usepackage{underscore}
\usepackage[table]{xcolor}
\usepackage{amsmath}
\usepackage[most]{tcolorbox}
\usepackage{fvextra}

\usepackage{parskip}

\definecolor{codegray}{RGB}{248,248,248}
\definecolor{bordergray}{RGB}{210,210,210}

\definecolor{boxback}{HTML}{F7F7F7}
\definecolor{boxborder}{HTML}{111111}

\newcolumntype{Y}{>{\raggedright\arraybackslash}X}
\newcolumntype{L}[1]{>{\raggedright\arraybackslash}p{#1}}

\newcommand{\tool}[1]{\texttt{\footnotesize #1}}

\lstdefinestyle{jsonstyle}{
  basicstyle=\ttfamily\footnotesize,
  columns=fullflexible,
  keepspaces=true,
  showstringspaces=false,
  breaklines=true,
  breakatwhitespace=false,
  frame=none,
  upquote=true
}

\newtcblisting{scenariojson}[2][]{
  enhanced,
  breakable,
  listing only,
  listing engine=listings,
  listing options={style=jsonstyle},
    colback=gray!3,
coltitle=black,
  colframe=gray!45,
  title={#2},
  fonttitle=\bfseries,
  boxrule=0.6pt,
  left=1mm,
  right=1mm,
  top=1mm,
  bottom=1mm,
  before skip=0.8em,
  after skip=0.8em,
  #1
}

\newtcolorbox{paperbox}[2][]{
  enhanced,
  breakable,
  coltitle=black,
  title={#2},
  fonttitle=\bfseries,
  sharp corners,
  boxrule=0.6pt,
  left=1.5mm,
  right=1.5mm,
  top=1mm,
  bottom=1mm,
  before skip=0.8em,
  after skip=0.8em,
  #1
}

\title{Synthetic Scenario Generation for Evaluation of Industry 4.0 Agents}

\author{
  Sagar Chethan Kumar \\
  \texttt{sagar.chethankumar@columbia.edu} \\
  Columbia University \\
  New York, NY, USA
  \And
  Rohith Kanathur \\
  \texttt{rk3443@columbia.edu} \\
  Columbia University \\
  New York, NY, USA
  \And
  Dhaval Patel \\
  \texttt{pateldha@us.ibm.com} \\
  IBM Research \\
  Yorktown Heights, NY, USA
    \And
  Kaoutar El Maghraoui \\
  \texttt{kelmaghr@us.ibm.com} \\
  IBM Research \\
  Yorktown Heights, NY, USA
}

\newtcolorbox{promptbox}[2][]{
  breakable,
  enhanced,
  colback=gray!3,
coltitle=black,
  colframe=gray!45,
  title={#2},
  fonttitle=\bfseries,
  boxrule=0.5pt,
  arc=2mm,
  left=2mm,
  right=2mm,
  top=1mm,
  bottom=1mm,
  #1
}

\begin{document}

\maketitle

\begin{abstract}
Industrial agent benchmarks require realistic evaluation scenarios that integrate telemetry, failure modes, maintenance records, and domain standards. However, existing benchmarks such as AssetOpsBench rely on manually authored scenarios and cover a limited set of asset classes. We extend AssetOpsBench with a Smart Grid Transformer asset class and four IEC-grounded diagnostic tools for health-index prediction, dissolved-gas analysis, winding-temperature assessment, and load-profile assessment. We further introduce \textsc{ScenarioGeneratorAgent}, a pipeline for synthetic industrial-agent scenario generation. The pipeline constructs evidence-grounded asset profiles, allocates coverage-aware scenario budgets across operational domains, and generates candidates through a hybrid validation-and-repair loop that enforces schema validity, tool reachability, physical plausibility, standards alignment, and deduplication. To improve scalability, we apply two-level caching, parallel focus-group generation, thread-pool offloading, batched LLM calls, and early rejection filtering. On Smart Grid Transformer scenario generation, these optimizations reduce end-to-end runtime by $8\times$ for 50 scenarios while preserving quality, achieving a composite quality score of $74.2 \pm 1.9$ compared with $73.8 \pm 3.0$ for the unoptimized baseline. These results show that standards-grounded synthetic scenario generation can efficiently expand industrial-agent benchmarks without sacrificing scenario quality.
\end{abstract}

\section{Introduction}

Industrial assets such as data-center cooling systems, power transformers, and wind turbines are
complex, multi-component systems whose safe and efficient operation depends on continuous monitoring
across heterogeneous data streams: time-series telemetry from IoT sensors, failure-mode libraries,
maintenance work orders, and operational logs.
Translating this flood of multi-modal data into timely, condition-aware decisions has historically required
scarce subject-matter experts (SMEs) who must synthesise insights across data by hand.
The emergence of large language model (LLM) based AI agents offers a qualitatively different
paradigm: autonomous, goal-driven systems that can integrate perception, reasoning, and action across
the full asset lifecycle, from anomaly detection and root-cause analysis to predictive maintenance
planning and work-order generation.

\texttt{AssetOpsBench}~\cite{patel2026assetopsbenchbenchmarkingaiagents} is the first unified framework designed to guide
the development, orchestration, and evaluation of domain-specific agents for Industry 4.0 asset
operations.
It provides a simulated industrial environment backed by a CouchDB IoT telemetry store, five
specialised agents (IoT, FMSR, TSFM, WO and Vibration), and a curated set of 141 manually authored scenarios
spanning single-agent utterances and multi-agent coordination tasks.

Despite this progress, two fundamental bottlenecks limit the benchmark's scope.
First, \textbf{asset-class coverage} is narrow: the initial release covers only chillers and air-handling
units (AHUs), both HVAC assets with well-understood failure modes.
Critical infrastructure classes, most prominently high-voltage power transformers whose health
assessment relies on distinct diagnostic modalities such as dissolved gas analysis (DGA) of insulating
oil, winding temperature monitoring, and electrical load profiling are entirely absent.
Agents evaluated on the existing benchmark cannot therefore be assessed on transformer diagnostic
tasks, which represent some of the most economically significant maintenance operations in power
utilities.
Second, and more fundamentally, \textbf{scenario creation does not scale}.
All 141 benchmark scenarios were manually authored by SMEs.
This process is time-consuming, difficult to standardise across asset classes, and requires deep domain
knowledge that is not readily available for every new asset type.
As AssetOpsBench is extended to new industrial domains, manual authoring becomes an acute
bottleneck. Each new asset class demands its own set of physically plausible, causally consistent,
tool-reachable scenarios that comply with relevant ISO and IEC standards.

These two gaps are compounded by a shared underlying difficulty: generating high-quality evaluation
scenarios for industrial agents is difficult because it requires satisfying simultaneous constraints
that are not all of the same kind.
A scenario must be \emph{physically plausible} (respecting the laws governing the asset), \emph{causally
consistent} (observations must lead to feasible actions), \emph{tool-reachable}
(all required MCP tools must be available with the right parameters), \emph{standards-compliant} (aligned
with ISO IEC standards), and \emph{non-repetitive} with respect to
existing scenarios.
Prompt-based generation alone is insufficient because an LLM may produce
scenarios that violate physical or operational constraints. Hence, a separate validation and repair stage is unavoidable.

This paper addresses both gaps  by making three contributions. First, we extend AssetOpsBench with a Smart Grid Transformer asset class, including MCP tools for health-index prediction, dissolved-gas analysis, winding-temperature assessment, and load-profile assessment. Second, we introduce ScenarioGeneratorAgent, a synthetic scenario generation pipeline that grounds scenario creation in live asset metadata, tool schemas, and retrieved domain evidence. Third, we evaluate the pipeline’s runtime and scenario quality, showing that caching, batching, thread-pool offloading, and parallel focus-group generation reduce generation time while preserving quality under static checks, LLM judging, and dry-run execution.

\section{Related Work}

\paragraph{Industrial asset management benchmarks.}
The original \texttt{AssetOpsBench}~\cite{patel2026assetopsbenchbenchmarkingaiagents}
is the closest prior work to ours.
It introduces a benchmark for industrial asset operations with manually authored scenarios,
specialised IoT, FMSR, TSFM, and WO agents, and an LLM-as-judge evaluation protocol.
Related industrial-agent work studies language-agent access to IoT telemetry~\cite{rayfield-etal-2025-react},
CodeReAct-style work-order automation~\cite{Zhou_Patel_Bhattacharyya_2026}, and
failure-mode/sensor reasoning~\cite{constantinides2025failuresensoriqmultichoiceqadataset}.
Other industrial benchmarks, such as SustainDC~\cite{naug2025sustaindcbenchmarkingsustainabledata}
and WFCRL~\cite{monroc2025wfcrlmultiagentreinforcementlearning}, evaluate control policies for
data centers and wind farms.
However, these efforts either rely on fixed expert-authored scenarios or focus on control and
diagnostic tasks rather than scalable scenario generation for new asset classes.
Our work extends \texttt{AssetOpsBench} by adding a new transformer asset class and replacing
manual scenario authoring with a standards-grounded generation pipeline.

\paragraph{Domain-specific and application-specific agent benchmarks.}
Several recent benchmarks evaluate LLM agents in specialised operational domains.
ITBench~\cite{jha2025itbenchevaluatingaiagents} evaluates agents on real-world IT automation
tasks spanning SRE, compliance, and FinOps, while AIOpsLab~\cite{chen2025aiopslabholisticframeworkevaluate}
constructs cloud-operations incidents from workload and fault generators.
Business, workplace, and software-engineering environments such as
CRMArena-Pro~\cite{huang2025crmarenaproholisticassessmentllm},
TheAgentCompany~\cite{xu2026theagentcompany}, and
SWE-bench~\cite{jimenez2024swebenchlanguagemodelsresolve} similarly test long-horizon
tool use in realistic digital workflows.
These benchmarks provide valuable execution-grounded evaluation protocols, but they do not
address the hybrid time-series, work-order, failure-mode, and standards-constrained setting of
industrial asset management.

\paragraph{Generalist and multi-agent benchmarks.}
Broader agent benchmarks evaluate general-purpose agents across web, OS, ML, and app-based
environments.
AgentBench~\cite{liu2025agentbenchevaluatingllmsagents}, WebArena~\cite{zhou2024webarenarealisticwebenvironment},
WorkArena~\cite{drouin2024workarenacapablewebagents}, OSWorld~\cite{xie2024osworldbenchmarkingmultimodalagents},
and AppWorld~\cite{trivedi2024appworldcontrollableworldapps} focus on interactive digital
environments, while MLEBench~\cite{chan2025mlebench}, MLAgentBench~\cite{huang2024benchmarking},
and MLGym~\cite{nathani2025mlgymnewframeworkbenchmark} evaluate agents on machine-learning
engineering and research workflows.
These settings are useful for studying planning, tool use, and execution reliability, but they
largely omit the physical-state reasoning and time-series modalities central to condition monitoring.
Our benchmark extension instead targets multi-agent reasoning over industrial telemetry,
diagnostic standards, maintenance evidence, and MCP-backed tools.

\paragraph{Automated and synthetic scenario generation.}
The difficulty of scaling handcrafted benchmarks has motivated automated task and environment
generation.
TaskBench~\cite{shen2024taskbenchbenchmarkinglargelanguage} constructs task-automation
benchmarks from tool graphs, AgentGen~\cite{hu2025agentgenenhancingplanningabilities}
synthesises planning environments and tasks, and ARE~\cite{froger2025arescalingagentenvironments}
scales agent environments with rules, tools, and verifiers.
APIGen~\cite{liu2024apigenautomatedpipelinegenerating}, ToolLLM~\cite{qin2023toolllmfacilitatinglargelanguage},
and WizardLM~\cite{xu2025wizardlmempoweringlargepretrained} further show that synthetic
instructions and function-calling data can improve tool-use capabilities.
However, these methods target general-purpose tool use rather than industrial scenarios requiring
physical plausibility, temporal consistency, standards alignment, and tool reachability.
Our pipeline generates scenarios from live MCP-exposed asset evidence and validates them through
a hybrid repair loop against engineering constraints, domain standards, and executable tool
availability.

\paragraph{Agent architectures and evaluation methodology.}
Our system builds on common agentic execution patterns, including ReAct-style reasoning and
acting~\cite{yao2023react}, executable code actions~\cite{wang2024executablecodeactionselicit},
plan-execute decomposition~\cite{xu2023rewoodecouplingreasoningobservations}, and
multi-agent planning~\cite{li2025agentoriented}.
Generalist multi-agent systems such as Magentic-One~\cite{fourney2024magenticonegeneralistmultiagentsolving}
motivate modular orchestration, while MCP-Universe~\cite{luo2025mcpuniversebenchmarkinglargelanguage}
and MCP-Bench~\cite{wang2025mcpbenchbenchmarkingtoolusingllm} evaluate agents over real MCP
servers.
For evaluation, we follow prior work on LLM-as-judge protocols~\cite{zheng2023judgingllmasajudgemtbenchchatbot}
and agent-as-judge evaluation~\cite{zhuge2024agentasajudgeevaluateagentsagents}, while using
trajectory-level failure analysis inspired by~\cite{cemri2026why}.
Unlike prior agent benchmarks, our evaluation setting couples trajectory quality with industrial
scenario validity, standards compliance, and executable tool solvability.

\paragraph{LLM-based diagnostic tools for power transformers.}
Transformer health assessment is a mature engineering discipline grounded in IEC 60599 (dissolved
gas analysis and Rogers Ratio fault classification) and IEC 60076-7 (thermal loading guidelines).
Prior ML work has applied neural networks to transformer fault diagnosis~\cite{rokani2023} and
health index prediction~\cite{velasquez2020,gorgan2010}, but these models have not been integrated
into an agent benchmarking framework or exposed as MCP tools.
Our \texttt{predict\_health\_index} and three LLM-based diagnostic tools (\texttt{interpret\_dga},
\texttt{assess\_winding\_temperature}, \texttt{assess\_load\_profile}) bridge this gap, providing the
first standards-grounded transformer diagnostic tooling within an open agent benchmark.


\section{Methodology}

\subsection{Transformer Asset Integration}

\begin{table}[h!]
\centering
\small
\caption{Smart Grid Transformer diagnostic tools added to AssetOpsBench.}
\label{tab:transformer-tools}
\begin{tabularx}{\linewidth}{@{}L{0.24\linewidth}YYY@{}}
\toprule
\textbf{Tool} & \textbf{Inputs} & \textbf{Output} & \textbf{Grounding} \\
\midrule

\tool{predict\_\allowbreak health\_\allowbreak index}
&
Oil, gas, and electrical, including dissolved gases, dielectric rigidity, power factor, and water content.
&
Transformer health score on a 0--100 scale and a discrete condition label.
&
Supervised random forest model trained on labeled transformer health-index data.
\\

\addlinespace

\tool{interpret\_\allowbreak dga}
&
Dissolved gas concentrations, including hydrogen, methane, acetylene, ethylene, and ethane.
&
Fault classification, Rogers Ratio codes, confidence level, and diagnostic reasoning.
&
IEC 60599 dissolved-gas-analysis interpretation and Rogers Ratio fault classification.
\\

\addlinespace

\tool{assess\_\allowbreak winding\_\allowbreak temperature}
&
Winding temperature, oil temperature, ambient temperature, alarm state, and trip state.
&
Thermal status, hot-spot rise, ageing rate, alarm/trip interpretation, and risk level.
&
IEC 60076-7 thermal loading and insulation ageing guidelines.
\\

\addlinespace

\tool{assess\_\allowbreak load\_\allowbreak profile}
&
Phase voltages, phase currents, line-to-line voltages, neutral current, and rated MVA.
&
Apparent load, load factor, current imbalance, neutral-current flag, and loading status.
&
IEC 60076-7 loading limits and three-phase apparent-power calculations.
\\

\bottomrule
\end{tabularx}
\end{table}

AssetOpsBench is extended with Smart Grid Transformers as a new asset
class, covering four signal families: health index prediction, dissolved gas analysis (DGA) of
transformer oil, winding temperature, and electrical load profiles.
The IoT agent automatically discovers the transformer assets and sensor readings.

Four new MCP tools are added to the FMSR agent.
\texttt{predict\_health\_index} uses a supervised random forest model trained on a
labeled transformer health dataset~\cite{velasquez2020} to predict a health
score from 0--100.
The remaining three tools are grounded in established electrical
engineering standards: \texttt{interpret\_dga} applies the IEC~60599
Rogers Ratio method~\cite{iec60599, Taha2015ImprovementOR} to classify
fault type from dissolved gas concentrations;
\texttt{assess\_winding\_temperature} applies the IEC~60076-7 thermal
model~\cite{iec60076} to compute insulation ageing rate and thermal
risk; and \texttt{assess\_load\_profile} derives three-phase apparent
load and classifies loading status against IEC~60076-7 loading
limits~\cite{iec60076, chapman2005electric}.
Grounding diagnostic tools in IEC standards makes their outputs easier to validate and ensures that generated scenarios can be checked against explicit engineering criteria.

\newpage

\subsection{Scenario Generation Pipeline}

Scaling \texttt{AssetOpsBench} to new asset classes remains bottlenecked by manual scenario authoring. Expert-written scenarios provide strong operational fidelity, but they are costly to produce, difficult to scale, and slow to adapt across heterogeneous assets, failure modes, sensors, and maintenance workflows.

\begin{figure}[H]
    \centering
    \includegraphics[width=1\linewidth]{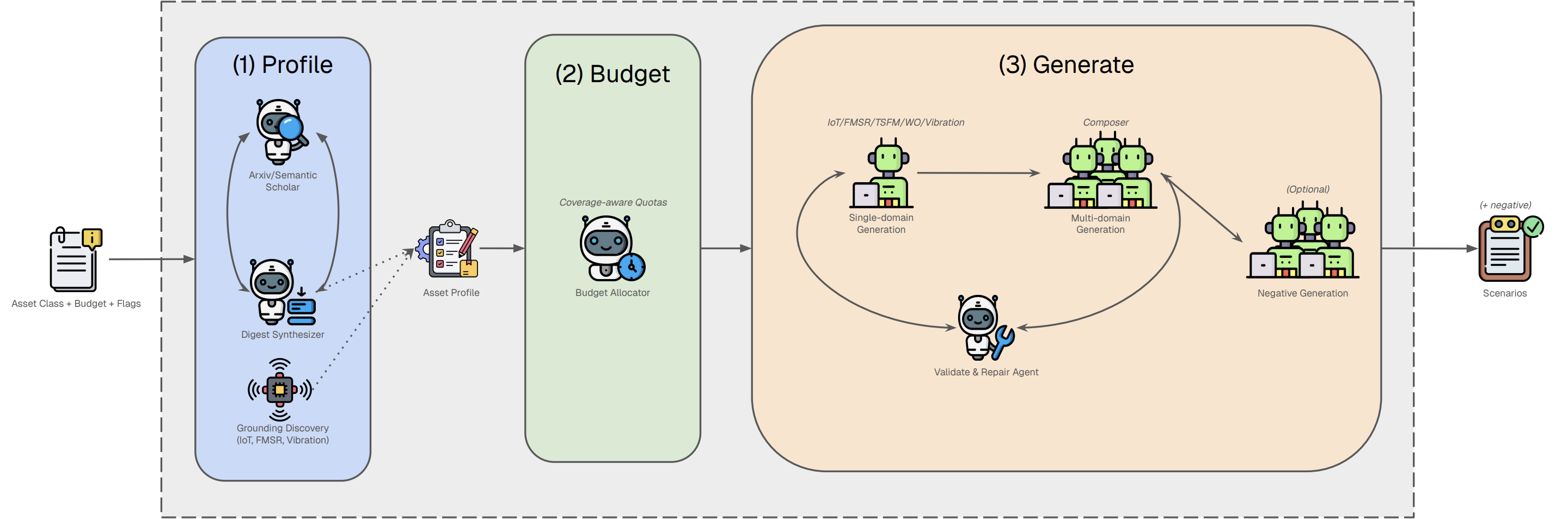}
    \caption{Scenario Generation Pipeline. The pipeline profiles the target asset, allocates a coverage-aware scenario budget, and generates validated positive and optional negative scenarios.}
    \label{fig:pipeline}
\end{figure}

We introduce a synthetic scenario generation pipeline that reduces this dependence on subject-matter experts while preserving grounding, diversity, and operational realism. As shown in Figure~\ref{fig:pipeline}, the pipeline proceeds in three stages: \textbf{profile}, \textbf{budget}, and \textbf{generate}. It first constructs an evidence-grounded asset profile, then allocates scenarios across operational domains, and finally synthesizes and validates scenarios under domain-specific constraints.

\subsubsection{Stage 1: Asset Profiling}

The first stage constructs an asset profile used for downstream scenario generation. Given the name of an asset class, we recover the operational context required to generate realistic operator- and manager-facing tasks. This includes the asset's role, major components, sensor modalities, observable degradation signals, failure modes, maintenance practices, relevant standards, and available tool interfaces. Rather than generating scenarios from an asset name alone, the pipeline grounds generation in a combination of live environment metadata, tool schemas, and curated external evidence.

To obtain domain evidence, we design a research agent that queries academic literature for asset-specific information. The agent performs query expansion over a fixed set of evidence categories derived from the scenario-design process used in AssetOpsBench~\cite{patel2026assetopsbenchbenchmarkingaiagents}. These categories are:

\begin{itemize}
    \item Condition monitoring, diagnostics, and degradation indicators.
    \item Maintenance practices, scheduling, and work-order-relevant context.
    \item Sensor modalities, measurements, and signal interpretation.
    \item Failure modes, faults, and operational risks.
    \item Applicable standards, norms, and industry conventions.
    \item Operator and manager-style tasks expressed in natural language.
    \item Gaps, limitations, or unsupported capabilities in the available evidence.
\end{itemize}

For each category, the research agent generates asset-specific search queries and retrieves the top-$k$ papers from scholarly sources (either Semantic Scholar or arXiv). Each retrieved paper is converted into a category-specific digest that extracts only the information relevant to the asset profile. The papers are then ranked by relevance, and the resulting digests are passed to a summarization and merging stage. This stage produces a coherent evidence brief that is later used to construct the final asset profile.

The profiling stage also supports \emph{grounding discovery}. In addition to external literature, the system inspects the execution environment to identify deterministically available entities and tool-accessible data. In AssetOpsBench, this includes CouchDB-backed information such as asset identifiers, sites, sensor tags, available time ranges and failure-mode mappings. This discovered grounding is used both to constrain scenario generation and to improve the research agent's queries. For example, if live data indicates that a transformer asset exposes dissolved-gas, winding-temperature, and load-current measurements, the research agent can prioritize literature on those modalities and their associated degradation mechanisms.

We distinguish between two generation regimes. In the \emph{closed-form} regime, each scenario is self-contained: the query includes all information required for evaluation and does not depend on live sensor data or environment-specific records. For example, a closed-form scenario may provide synthetic measurements directly in the user query and ask the agent to assess whether the asset exhibits signs of abnormal operation. In contrast, in the \emph{open-form} regime, scenarios reference concrete entities in the execution environment, such as asset IDs, sensor tags, time ranges, alerts, or work-order records. Open-form scenarios are generated only when these entities can be deterministically resolved from the available environment.

The resulting asset profile therefore serves as the grounding interface between domain evidence, live data inventories, and scenario generation. It ensures that generated scenarios align with observable asset signals, known failure mechanisms, available tools, and the concrete execution context.

\subsubsection{Stage 2: Domain Budgeting}

The second stage allocates the target scenario count across operational domains. Given an asset profile, the budgeter assigns generation quotas to domain agents, including IoT, FMSR, TSFM, WO, and Vibration when supported. Budgets may be specified manually, but by default they are derived from the asset's available tools, data modalities, failure-mode coverage, and maintenance context.

This stage prevents over-representation of domains with abundant surface-level evidence while preserving coverage of operationally important capabilities. For example, assets with rich telemetry are assigned more monitoring and time-series scenarios, whereas assets with maintenance records or lifecycle information receive greater work-order and decision-support coverage. Domains without sufficient grounding, such as vibration analysis for assets without vibration sensors, are downweighted or excluded.

Multi-agent scenarios are budgeted separately to ensure that cross-domain coordination is represented without dominating the benchmark. The output of this stage is a domain-level scenario budget used to control coverage during generation.

\subsubsection{Stage 3: Scenario Generation and Validation}

The third stage generates candidate scenarios conditioned on the asset profile, domain budget, and domain-specific policies. Each scenario follows the AssetOpsBench scenario schema, consisting of an identifier, task type, natural-language query, operational category, and characteristic form. When required, scenarios also include grounding metadata linking the task to concrete environment entities.

\begin{scenariojson}[label={box:example-scenario}]{Example text and characteristic form for a generated scenario.}
{
  "text": "For Transformer 1 at MAIN, compare CO and CO2 levels between 2024-01-15 and 2024-01-19. If CO2 data is missing, use oxygen levels instead. Summarize the comparison and predict the next work-order type from historical data.",
  "characteristic_form": "A correct response retrieves gas histories using get_history, compares CO and CO2 over the requested interval, applies the oxygen fallback if CO2 is unavailable, and then calls predict_next_work_order to forecast the next work-order type."
}
\end{scenariojson}

Generation is constrained by domain policies that define valid tools, task categories, grounding patterns, and prohibited constructions. These policies ensure that telemetry scenarios reference valid sensors and time ranges, failure-mode scenarios preserve plausible component-degradation relationships, and work-order scenarios align with maintenance-oriented actions.

Each candidate is passed through a bounded validation and repair loop. An LLM-based repair step corrects malformed or underspecified candidates, while deterministic checks enforce schema validity, tool-domain consistency, grounding validity, duplicate removal, and budget compliance. Candidates that remain invalid after bounded retries are discarded.

Multi-agent scenarios are constructed by composing validated single-domain scenarios into cross-domain tasks. These scenarios require agents to integrate information across operational views, such as connecting sensor anomalies to failure modes or maintenance history. The composed scenarios are revalidated before inclusion.

The pipeline also supports optional negative scenarios. These are generated after the positive set and do not consume the domain budget. Negative scenarios introduce invalid identifiers, unsupported dependencies, missing time windows, or insufficient evidence, enabling evaluation of refusal behavior and uncertainty calibration.

\subsection{Model and Optimization Scope}

The \texttt{ScenarioGeneratorAgent} pipeline uses
\texttt{watsonx/meta-llama/llama-3-3-70b-instruct} via IBM WatsonX
as the LLM backbone.
The model is invoked across all pipeline stages: asset profiling,
budget allocation, and scenario generation. 

The unoptimized baseline executes asset profiling, literature retrieval, budget allocation, and scenario generation sequentially, without persistent caching, batched scenario generation, early rejection, or thread-pool offloading. The optimized variant enables all five optimizations described below.

\begin{itemize}
    \item \textbf{Two-level caching.} Asset profiles, retrieved literature, and few-shot examples are cached at two levels: a fast in-memory LRU cache and persistent disk storage. This avoids re-running expensive LLM synthesis and academic retrieval for the same asset across repeated runs.

    \item \textbf{Parallel per-domain generation.} Scenario generation across focus groups (IoT, FMSR, TSFM, Work Order) runs parallelly rather than sequentially. 
    
    \item \textbf{Thread pool offloading.} Blocking LLM HTTP calls are offloaded to a shared thread pool executor, keeping the async event loop free and allowing multiple pipeline phases to make progress simultaneously without stalling on I/O.

    \item \textbf{Batch LLM calls.} Multiple scenarios are requested in a single prompt rather than one call per scenario, amortizing per-call overhead, network latency, parsing, and rate-limiter tokens.

    \item \textbf{Early rejection filtering.} A lightweight pre-screen eliminates clearly malformed or duplicate candidates before the full LLM validator runs, avoiding unnecessary validation cost for outputs that fail basic length, field-presence, or near-duplicate checks.
\end{itemize}


\section{Experimental Results}
\label{sec:experiments}

We conduct four profiling experiments to characterize the latency behavior of the scenario generator agent pipeline and to quantify the impact of the optimization techniques applied.
All experiments use the Smart Grid Transformer asset in  CouchDB-grounded generation mode. Times are reported in wall-clock seconds. Unless specified, all experiments use the following defaults: batch size of 10 scenarios per LLM call, 3 focus groups scenario generation in parallel, 4 threads in the I/O thread pool, a 24-hour disk cache for asset profiles, CouchDB grounded scenario generation, and ArXiv as the literature retriever. All measurements are averaged over 3 runs.
 
\subsection{Experiment 1: Scalability -- Pipeline Time vs.\ Number of Scenarios}

We vary $N \in \{10, 25, 50\}$ while holding all other parameters at
their defaults, running independently on both the unoptimised baseline
and the optimised variant to measure how end-to-end latency scales with
workload size.

\begin{table}[H]
\centering
\caption{Pipeline time vs.\ number of scenarios.}
\label{tab:scalability}
\small
\begin{tabular}{lcccccc}
\toprule
& \multicolumn{3}{c}{\textbf{Baseline}} & \multicolumn{3}{c}{\textbf{Optimised}} \\
\cmidrule(lr){2-4} \cmidrule(lr){5-7}
\textbf{Phase} & \textbf{10} & \textbf{25} & \textbf{50}
               & \textbf{10} & \textbf{25} & \textbf{50} \\
\midrule
Get Server Descriptions         & 2.40 s   & 2.58 s   & 2.20 s  & 1.56 s  & 2.42 s  & 2.16 s \\
Build Asset Profile             & 448.28 s & 338.90 s & 325.70 s & \textbf{0.00 s}  & \textbf{0.00 s} & \textbf{0.00 s}  \\
Allocate Scenario Budget        & 2.05 s  & 3.37 s  & 2.34 s  & \textbf{0.00 s} & \textbf{0.00 s} & \textbf{0.00 s} \\
Generate \& Validate Single     & 11.40 s & 34.79 s & 38.83 s & \textbf{8.23 s} & \textbf{14.86 s} & \textbf{15.95 s} \\
Generate \& Validate Multi      & 4.05 s  & 20.67 s & 39.32 s & 10.73 s & 17.28 s & 18.11 s \\
\midrule
\textbf{Full Pipeline (in seconds)}                  & 468.22 s & 400.35 s & 408.43 s
                                & \textbf{13.63 s}  & \textbf{19.88 s}  & \textbf{50.86 s} \\
\bottomrule
\end{tabular}
\end{table}

In the baseline, Build Asset Profile dominates at 325--448\,s
(80--96\% of total time) due to uncached literature retrieval and LLM
synthesis on every run.
The optimised variant eliminates this entirely via a two-level cache
(L1 in-memory LRU + L2 disk), reducing it to zero on warm runs and
delivering speedups of \textbf{34$\times$} at $N=10$, \textbf{20$\times$}
at $N=25$, and \textbf{8$\times$} at $N=50$.
The decreasing speedup ratio at larger $N$ reflects the growing share of
generation time, which scales linearly with budget (Phase~3:
8.2\,s $\to$ 15.9\,s) but remains well below the baseline (38.8\,s
at $N=50$), attributable to parallel focus-group execution.

\subsection{Experiment 2: Parallelism -- Pipeline Time vs.\ Number of Parallel Focuses}

In the optimized variant, we vary \texttt{C} $\in \{1, 2, 3, 5\}$, which controls how
many of the five focus-group generation tasks (\texttt{iot}, \texttt{fmsr},
\texttt{tsfm}, \texttt{wo}, \texttt{vibration}) execute in parallel.
All other parameters are fixed at $N=50$; the asset profile and budget are served from disk cache so Phases~1 and~2 contribute near zero time.
 
\begin{table}[H]
\centering
\caption{Pipeline time vs.\ number of parallel focuses.}
\label{tab:parallelism}
\small
\begin{tabular}{lcccc}
\toprule
\textbf{Phase} & \textbf{C=1} & \textbf{2} & \textbf{3} & \textbf{5} \\
\midrule
Get Server Descriptions           &  2.18 s &  2.15 s &  2.18 s &  2.45 s \\
Build Asset Profile               &  \textbf{0.00 s} &  \textbf{0.00 s} &  \textbf{0.00 s} &  \textbf{0.00 s} \\
Allocate Scenario Budget          &  \textbf{0.00 s} &  \textbf{0.00 s} &  \textbf{0.00 s} &  \textbf{0.00 s} \\
Gen Single-Agent & \textbf{43.88 s} & \textbf{22.29 s} & \textbf{17.66 s} & \textbf{17.62 s} \\
Gen Multi-Agent  & 46.07 s & 54.51 s & 44.44 s & 27.22 s \\
\midrule
\textbf{Full Pipeline (in seconds)}             & \textbf{87.39 s} & \textbf{78.95 s} & \textbf{64.29 s} & \textbf{47.30 s} \\
\bottomrule
\end{tabular}
\end{table}

Increasing \texttt{C} from 1 (fully serial) to 5 reduces total pipeline time from 87.4\,s to 47.3\,s, a \textbf{1.85$\times$} speedup.
Phase~3 exhibits the clearest scaling: 43.9\,s (C=1) $\to$ 17.6\,s (C=5), with sharply diminishing returns beyond 3 concurrent tasks.
 
\subsection{Experiment 3: Caching -- Impact of Cache State on Pipeline Time}

We compare two runs of the optimized variant with identical parameters
($N=50$, all defaults) differing only in cache state.
The \emph{cold} run forces a full asset profile; the \emph{warm} run reads the profile written by the cold run.
 
\begin{table}[H]
\centering
\caption{Pipeline time vs.\ cache state.}
\label{tab:caching}
\small
\begin{tabular}{lcc}
\toprule
\textbf{Phase} & \textbf{Cache Cold} & \textbf{Cache Warm} \\
\midrule
Get Server Descriptions           &   2.41 s &   2.28 s \\
Build Asset Profile               & 247.40 s &   \textbf{0.00 s} \\
Allocate Scenario Budget          &  \textbf{0.00 s} &   \textbf{0.00 s} \\
Generate \& Validate Single-Agent &  17.30 s &  17.23 s \\
Generate \& Validate Multi-Agent  &  35.34 s &  45.57 s \\
\midrule
\textbf{Full Pipeline (in seconds)}             & \textbf{302.47 s} & \textbf{65.09 s} \\
\bottomrule
\end{tabular}
\end{table}

The disk cache eliminates Phase~1 entirely on warm runs (247.4\,s $\to$
0\,s), saving over four minutes per run.
The full pipeline drops from 302.5\,s (cold) to 65.1\,s (warm): a
\textbf{4.6$\times$} speedup.
The small increase in Phase~4 (35.3\,s $\to$ \ 45.6\,s) falls
within normal LLM API latency variance.
 
\subsection{Experiment 4: Combined Optimizations -- Baseline vs.\ Optimized (Cold Cache)}
 
To fairly isolate the non-caching optimizations, we compare the unoptimized baseline against the optimized variant under identical cold-cache conditions
 
\begin{table}[H]
\centering
\caption{Pipeline time: baseline vs.\ optimized, both cold cache.}
\label{tab:combined}
\small
\begin{tabular}{lcc}
\toprule
\textbf{Phase} & \textbf{Baseline} & \textbf{Optimized} \\
\midrule
Get Server Descriptions           &   2.20 s &   2.42 s \\
Build Asset Profile               & \textbf{325.70 s} & \textbf{216.85 s} \\
Allocate Scenario Budget          &   2.34 s &   2.56 s \\
Generate \& Validate Single-Agent &  \textbf{38.83 s} &  \textbf{14.65 s} \\
Generate \& Validate Multi-Agent  &  39.32 s &  32.46 s \\
\midrule
\textbf{Full Pipeline (in seconds)}             & \textbf{408.43 s} & \textbf{268.96 s} \\
\bottomrule
\end{tabular}
\end{table}

Even with the cache disabled, the optimized variant achieves a
\textbf{1.52$\times$} end-to-end speedup (408.4\,s $\to$ 269.0\,s).
Phase~1 improves by 33\% (325.7\,s $\to$ 216.9\,s) through I/O concurrency: the baseline executes PDF downloads, database queries, and literature searches sequentially, whereas the optimized module offloads each blocking call to a background thread pool, filling CPU-idle network-wait time with concurrent requests.
Phase~3 shows the largest relative gain: a \textbf{2.65$\times$} speedup (38.8\,s $\to$ 14.7\,s) from parallel focus execution.

\section{Evaluation}
 
Optimizing the generation pipeline for speed is only meaningful if the
quality of the produced scenarios is preserved. A faster pipeline that
generates low-quality, unanswerable, or structurally broken scenarios
would be counterproductive for benchmarking purposes.
This section presents the evaluation methodology used to assess scenario
quality and a comparative quality analysis between the baseline and
optimized pipelines.
Scenario quality is assessed along three dimensions, summarized in
Table~\ref{tab:eval_methodology}, summing to a 0--100 composite score.
A scenario above 70 is considered high quality; below 50 indicates
fundamental issues. Dry-run execution carries the highest weight (50 pts) as it provides the strongest real-world signal: a scenario an agent can execute and answer plausibly is useful for benchmarking.
 
To verify that the optimization techniques do not degrade scenario quality,
50 Smart Grid Transformer scenarios were generated using both the unoptimized
baseline and the optimized pipeline under identical conditions (CouchDB-grounded, $N=50$), each repeated across three independent runs.
Both sets were evaluated using the full three-stage methodology.
Table~\ref{tab:quality} reports mean $\pm$ standard deviation across the
three runs for each component score and the composite quality score.
 
\begin{table}[H]
\centering
\caption{Scenario quality evaluation: baseline vs.\ optimized
 (mean $\pm$ std, $n=3$ runs).}
\label{tab:quality}
\small
\begin{tabular}{lcc}
\toprule
\textbf{Metric} & \textbf{Baseline} & \textbf{Optimized} \\
\midrule
Static Score (/20)       & $18.4 \pm 0.6$ & $19.1 \pm 0.3$ \\
LLM Judge Score (/30)    & $22.2 \pm 3.6$ & $22.4 \pm 1.8$ \\
Dry-Run Score (/50)      & $33.1 \pm 1.3$ & $32.7 \pm 2.5$ \\
\midrule
\textbf{Quality Score (/100)} & $\mathbf{73.8 \pm 3.0}$ & $\mathbf{74.2 \pm 1.9}$ \\
\bottomrule
\end{tabular}
\end{table}
 
The mean quality score of the optimized pipeline ($74.2 \pm 1.9$) is
statistically indistinguishable from the baseline ($73.8 \pm 3.0$), with a gap of only 0.4 points well within the run-to-run variance of either condition.
Both pipelines exceed the 70-point high-quality threshold on average,
confirming that the generated scenarios are suitable for agent benchmarking.

\section{Limitations and Future Work}

All experiments were conducted using a single LLM backend (WatsonX
Llama~3.3~70B), so it is unknown whether the speedup ratios and quality
scores hold across different model families.
Cross-model comparison was planned but could not be conducted due to model unavailability of closed source models; open-source alternatives would additionally require domain-specific fine-tuning to produce high quality scenarios.
Finally, the dry-run execution and quality judging are performed by an LLM, meaning the scores are inherently sensitive to the judge model's own capabilities and biases.

An extension to this work is cross-model comparison. Repeating the experiments across GPT-4, Llama~4, and Mistral Large would
establish whether pipeline throughput is sensitive to backend latency
characteristics.
Broader asset class coverage including wind turbines, pumps, compressors would validate that the generation and evaluation methodology generalizes beyond transformers.
On the evaluation side, replacing the LLM judge with human annotations is another future work worth looking into.

\section{Conclusion}
We have presented two contributions that together address the dual
bottleneck of asset-class coverage and scenario-authoring scalability in
\texttt{AssetOpsBench}.
First, we extended the benchmark with a complete Smart Grid Transformer
asset class containing four failure mode diagnostic tools spanning a
supervised health index predictor and three IEC standards-grounded
LLM-based diagnostic tools (Rogers Ratio DGA interpretation, winding temperature assessment, and load profile assessment).
This is, to our knowledge, the first integration of standards-grounded
transformer diagnostic tooling within an open source benchmark.
Second, we introduced a \emph{ScenarioGeneratorAgent} pipeline that
automates the construction of physically plausible,  tool-reachable, and standards-compliant evaluation scenarios for any
asset class on-boarded into the benchmark.
The pipeline combines grounded asset-profile synthesis, budgeted
domain allocation, focus-group generation, and a
validate-and-repair loop. To the best of our knowledge, this is the first work to apply automated,
knowledge-grounded scenario generation to industrial asset maintenance
operations.

Our study on the Smart Grid Transformer asset class scenario generation demonstrates key optimizations such as two-level caching, asynchronous I/O thread pooling, and parallel focus-group execution, that reduce end-to-end pipeline time by up to \textbf{8$\times$} for 50 scenarios over the unoptimized baseline, profiled using PyTorch Profiler and W\&B across experiments varying scenario budget, parallelism, and cache state.
Crucially, these gains do not come at the cost of scenario quality: the
optimized pipeline attains a composite quality score (on a 100-point scale) of $74.2 \pm 1.9$
versus $73.8 \pm 3.0$ for the baseline, a difference well within
run-to-run variance.

\section*{Acknowledgements}

We sincerely thank Dr.\ Dhaval Patel and Dr.\ Kaoutar El Maghraoui from IBM Research for their mentorship and guidance throughout this work, and for the opportunity to collaborate.

\newpage

\bibliographystyle{abbrvnat}
\bibliography{references}

\newpage

\appendix

\section{Evaluation Criteria for Scenarios}
\begin{table*}[h!]
\centering
\caption{Evaluation dimensions and scoring criteria.}
\label{tab:eval_methodology}
\small
\begin{tabular}{llp{11cm}}
\toprule
\textbf{Dimension} & \textbf{Points} & \textbf{Criteria} \\
\midrule
\multirow{5}{*}{Static checks} & \multirow{5}{*}{20}
  & Schema completeness (all required fields present) \\
& & Type diversity (no single type $>$50\% of dataset) \\
& & Category alignment (maps to a recognised problem group) \\
& & Text length (15--400 characters) \\
& & Duplicate detection (IDs and text) \\
\midrule
\multirow{5}{*}{LLM judge} & \multirow{5}{*}{30}
  & Answerability: can the task be answered? (10 pts) \\
& & Tool usability: are the required tools and parameters available? (8 pts) \\
& & Difficulty: not trivial, not impossibly complex (6 pts) \\
& & Characteristic quality: grading rubric is specific and measurable (4 pts) \\
& & Clarity: task is unambiguous to a domain expert (2 pts) \\
\midrule
\multirow{3}{*}{Dry-run execution} & \multirow{3}{*}{50}
  & Plausible response: final answer addresses the task (20 pts) \\
& & Correct tools used: fraction of steps using appropriate tools (15 pts) \\
& & No obvious errors: fraction of steps completing without error (15 pts) \\
\midrule
\textbf{Total} & \textbf{100} & \\
\bottomrule
\end{tabular}
\end{table*}

\begin{table}[H]
\centering
\caption{Roger ratio codes and fault diagnosis~\cite{Taha2015ImprovementOR}}
\label{tab:iec_codes}
\small
\begin{tabular}{llll}
\toprule
\textbf{Code (R\textsubscript{1}, R\textsubscript{2}, R\textsubscript{3})} &
\textbf{Fault Type} & \textbf{Code (R\textsubscript{1}, R\textsubscript{2}, R\textsubscript{3})} &
\textbf{Fault Type} \\
\midrule
(0, 0, 0)     & NF                 & (1, 0, 2)     & HED \\
(0, 1, 0)     & PDLED          & (0, 0, 1)     & LTH1\\
(1, 1, 0)     & PDHED         & (0, 2, 0)     & LTH4\\
(1 or 2, 0, 1 or 2) & LED  & (0, 2, 1)     & MTH\\
Other         & UD           & (0, 2, 2)     & HTH\\
\bottomrule
\end{tabular}
\end{table}

\section{Prompt Templates}
\label{app:prompts}

\begin{promptbox}{Asset Profile Enrichment}
\begin{Verbatim}[breaklines=true,breakanywhere=true,fontsize=\footnotesize]
You are an expert reliability engineer preparing study-backed enrichment for an Asset Profile used in scenario generation.

The system will compose the final Asset Profile in code. Your job is to return only the study-derived fields that are not already deterministically grounded from CouchDB discovery.

Asset Name: {asset_name}
Generation Mode: {generation_mode}

Grounded Coverage Summary:
{grounding_summary_json}

Research Digest:
{research_digest}

Available Subagent Tools:
{tool_descriptions}

Task:
Produce a JSON object matching the following schema exactly.
CRITICAL: Output ONLY the raw JSON object. Do NOT include markdown code blocks, Python code, or any conversational preamble.

Rules:
- Keep the response concise. Do NOT restate asset instances, asset_name, or generation_mode.
- In open_form mode, treat grounded live coverage as the source of truth for concrete site names, asset ids, sensor keys (listed under `iot_sensors` / `vibration_sensors` in the summary), per-asset time ranges, and timestamps.
- In closed_form mode, assume scenarios must be self-contained: the query text should list explicit sensor readings (sensor name, numeric value, unit), and may include rules or dataset names when relevant.
- Use the research digest as the primary source of truth for asset-class understanding, operator workflows, monitoring practices, and standards.
- Do NOT return `failure_sensor_mapping` or `sensor_failure_mapping`.
- Use `iot_sensors` and `vibration_sensors` as separate lists of objects, each with `name` and `description` (what it measures). The system merges these with sensor name strings from the grounding summary in code (every IoT name from the summary must appear under `iot_sensors`; every vibration name from the summary under `vibration_sensors`).
- Use `failure_modes` the same way as before: list of {{"key": "...", "description": "..."}}. The system merges with grounded failure modes and F2S/S2F mappings in code (mappings use keys only).
- Include multiple relevant tools per focus when the toolset clearly supports more than one important action for that focus.
- Use an empty array [] for `vibration_sensors` when vibration sensing does not apply; use [] for `iot_sensors` only when the asset truly has no IoT channels in the grounded summary.
- Only include sensor descriptions, failure mode descriptions, and task types that are clearly supported by the research digest, grounded coverage summary, or tool definitions.
- If the research digest is sparse, stay conservative rather than inventing highly specific details.
- The profile must represent the physical industrial asset class itself, not surrounding digital, networking, cyber, or facility infrastructure.
- Add realistic tasks from the point of view of industrial asset operator and the operator's manager.
- Include vibration tools only when vibration analysis is relevant to the asset class or grounded data; otherwise use [] for "vibration".

{{
    "description": "Short summary of the asset class (1-2 sentences).",
    "iot_sensors": [
        {{"name": "sensor_name", "description": "what it measures"}}
    ],
    "vibration_sensors": [
        {{"name": "sensor_name", "description": "what it measures"}}
    ],
    "failure_modes": [
        {{"key": "failure_mode_name", "description": "why it matters for this asset"}}
    ],
    "relevant_tools": {{
        "iot": [
            {{"name": "tool_name_a", "reason": "why this tool matters for the asset"}},
            {{"name": "tool_name_b", "reason": "why this tool matters for the asset"}}
        ],
        "fmsr": [
            {{"name": "tool_name_a", "reason": "why this tool matters for the asset"}},
            {{"name": "tool_name_b", "reason": "why this tool matters for the asset"}}
        ],
        "tsfm": [
            {{"name": "tool_name_a", "reason": "why this tool matters for the asset"}},
            {{"name": "tool_name_b", "reason": "why this tool matters for the asset"}}
        ],
        "wo": [
            {{"name": "tool_name_a", "reason": "why this tool matters for the asset"}},
            {{"name": "tool_name_b", "reason": "why this tool matters for the asset"}}
        ],
        "vibration": [
            {{"name": "tool_name_a", "reason": "why this tool matters for the asset"}},
            {{"name": "tool_name_b", "reason": "why this tool matters for the asset"}}
        ]
    }},
    "operator_tasks": ["task phrased from the operator point of view"],
    "manager_tasks": ["task phrased from the manager point of view"]
}}
\end{Verbatim}
\end{promptbox}

\begin{promptbox}{Scenario Budget Allocation}
\begin{Verbatim}[breaklines=true,breakanywhere=true,fontsize=\footnotesize]
You are an expert Scenario Strategy Consultant for AssetOps Bench.

Given an Asset Profile and a total number of scenarios to generate ({total_scenarios}), your task is to allocate this budget across the following six focus areas:
1. iot: Focusing on sensor data and basic telemetry.
2. fmsr: Focusing on failure modes and structural reliability.
3. tsfm: Focusing on time-series analysis and technical maintenance.
4. wo: Focusing on actual maintenance execution and work orders.
5. vibration: Focusing on vibration diagnostics, severity assessment, FFT/envelope workflows, and bearing-related reasoning.
6. multiagent: Complex, multi-stage workflows involving orchestration of multiple agents.

Asset Profile:
{asset_profile_json}

Allocation Strategy:
- Prioritize agents that have more "relevant_tools" or richer "failure_modes" / "iot_sensors" / "vibration_sensors" entries in the profile.
- If the asset mentions complex standards (ISO 14224, etc.), lean towards analytical categories (tsfm, fmsr, vibration where appropriate).
- It is perfectly acceptable for a category to have 0 scenarios if the Asset Profile doesn't warrant it (e.g the sensor/domain is irrelevant for an asset class).
- Cap "multiagent" at a maximum of 75% of the total budget (e.g., max 37 if total is 50).
- The sum of all allocations MUST exactly equal {total_scenarios}.

Output Format:
CRITICAL: Return ONLY a raw JSON object. Do NOT include markdown code blocks, Python code, or any conversational preamble.

{{
    "reasoning": "A brief explanation of why you chose this distribution based on the asset details.",
    "allocation": {{
        "iot": int,
        "fmsr": int,
        "tsfm": int,
        "wo": int,
        "vibration": int,
        "multiagent": int
    }}
}}
\end{Verbatim}
\end{promptbox}

\begin{promptbox}{Scenario Generation Prompt}
\begin{Verbatim}[breaklines=true,breakanywhere=true,fontsize=\footnotesize]
You are an advanced industrial agent Scenario Architect for AssetOps Bench.
We need you to generate {count} evaluation scenarios with primary focus '{subagent_name}' for the asset class: {asset_name}.

Generation Mode:
{generation_mode}

Asset Profile (full JSON):
{asset_profile_json}

Available Focus Tools:
{tool_definitions}

{few_shot_examples_section}

When generating scenarios, use the Generation Mode and Mode-specific grounding rules (not the few-shot rows) to decide whether each scenario must embed readings, rules, summaries, or dataset identifiers in the query text.

Suggested category values for this focus:
{category_options}

Primary-focus requirements:
{specialization_requirements}

Hard-scenario contract for this batch:
- At least {hard_target_count} of the {count} scenarios should satisfy the hard rubric.
- Hard rubric: ask for at least two distinct outputs or actions, include at least one explicit constraint, and include an if/else, fallback, or missing-data instruction.
- Focus-specific hard patterns:
{hardness_guidance}

Avoid these bad patterns:
{forbidden_patterns}

Mode-specific grounding rules:
{mode_requirements}

Already accepted scenario texts. Avoid duplicates or near-duplicates of any of them:
{accepted_scenario_texts}

Constraints:
1. Every scenario must read like a realistic direct request from an industrial operator or the operator's manager.
2. Prefer end-user-centric wording such as "Will my transformer's health be okay tomorrow?" over tool-centric or benchmark-centric wording such as "predict transformer health".
3. The scenario may involve supporting work from other agents, but the main burden should stay on the primary focus '{subagent_name}'.
4. Every scenario must be highly specific, having a clear 'text', a 'category', and a 'characteristic_form'.
5. The characteristic_form must explicitly mention the concrete MCP tool names needed to solve the task.
6. The 'text' field must stay natural operator language only: do not name MCP tools, API or function identifiers, or add parenthetical hints such as "e.g. get_failure_modes tool". Reserve every concrete tool name for 'characteristic_form' only.
7. Most scenarios should be multi-part and instruction-following rather than short one-liners.
8. Closed-form scenarios must embed explicit inline sensor readings in the query text: for each measurement, sensor name (or label), numeric value, and unit (e.g. ppm, %, Hz, mm/s). You may also embed rule text, summaries, or dataset identifiers when the task requires them.
9. Open-form scenarios must use only grounded identifiers present in the Asset Profile.
10. Do not output Unsupported.

Task: Generate a JSON array of {count} scenarios.
CRITICAL: Output ONLY the raw JSON array. Do NOT include markdown code blocks, Python code, or any conversational preamble.

Format exactly (raw JSON only):
[
    {{
        "text": "...",
        "category": "...",
        "characteristic_form": "..."
    }}
]
    \end{Verbatim}
\end{promptbox}

\section{Transformer Diagnostic Tools}
\label{app:tools}

\begin{promptbox}{Transformer Health index prediction}
\begin{Verbatim}
$ uv run plan-execute --show-plan --show-history "What is the health
  index of a transformer with the following sensor readings: hydrogen
  2845 ppm, oxygen 5860 ppm, nitrogen 27842 ppm, methane 7406 ppm,
  CO 32 ppm, CO2 1344 ppm, ethylene 16684 ppm, ethane 5467 ppm,
  acetylene 7 ppm, DBDS 19 ppm, power factor 1, interfacial voltage
  45, dielectric rigidity 55, and water content 0."
 
Plan
  [1] fmsr: Predict the health index of the transformer based on the
       given sensor readings.
       tool: predict_health_index
       args: {'asset_name': 'transformer', 'hydrogen': 2845,
              'oxygen': 5860, 'nitrogen': 27842, 'methane': 7406,
              'co': 32, 'co2': 1344, 'ethylene': 16684,
              'ethane': 5467, 'acetylene': 7, 'dbds': 19,
              'power_factor': 1, 'interfacial_v': 45,
              'dielectric_rigidity': 55, 'water_content': 0}
       deps=none | expected: The predicted health index of the
       transformer.
 
Execution History
  [OK] Step 1 (fmsr): Predict the health index of the transformer
       based on the given sensor readings.
       tool: predict_health_index
       args: {'asset_name': 'transformer', 'hydrogen': 2845, ...}
       {
         "asset_name": "transformer",
         "health_index": 72.38500000000008,
         "condition": "Good"
       }
 
Answer
  The health index of the transformer is 72.39, 
  indicating a "Good" condition.
\end{Verbatim}
\end{promptbox}

\begin{promptbox}{Transformer Interpret DGA}
\begin{Verbatim}
$ uv run plan-execute --show-plan --show-history "Interpret the DGA
  gas readings for transformer: Hydrogen 100 ppm, Methane 50 ppm,
  Acetylene 5 ppm, Ethylene 20 ppm, Ethane 10 ppm."
 
Plan
  [1] fmsr: Interpret the DGA gas readings for the transformer with
       given gas concentrations.
       tool: interpret_dga
       args: {'asset_name': 'transformer', 'hydrogen': 100,
              'methane': 50, 'acetylene': 5, 'ethylene': 20,
              'ethane': 10}
       deps=none | expected: Interpretation of the DGA gas readings
       for the transformer.
 
Execution History
  [OK] Step 1 (fmsr): Interpret the DGA gas readings for the
       transformer with given gas concentrations.
       tool: interpret_dga
       args: {'asset_name': 'transformer', 'hydrogen': 100,
              'methane': 50, 'acetylene': 5, 'ethylene': 20,
              'ethane': 10}
       {
         "asset_name": "transformer",
         "fault_type": "LED",
         "r1": 0.5,
         "r2": 0.25,
         "r3": 2.0,
         "code": "(1,0,1)",
         "confidence": "Medium",
         "reasoning": "The code combination (1,0,1) corresponds ..."
       }
 
Answer
  The DGA gas readings for the transformer indicate a fault type of
  Low-Energy Discharge (LED) with medium confidence, based on the
  IEC 60599 ratio method, suggesting a thorough inspection and
  maintenance of the transformer is recommended.
\end{Verbatim}
\end{promptbox}

\begin{promptbox}{Transformer Winding Temperature Assessment}
\begin{Verbatim}
$ uv run plan-execute --show-plan --show-history "Assess the winding
  temperature for transformer 'TX1' with WTI 80, OTI 85, ATI 78,
  OTI_A 2, OTI_T 1."
 
Plan
  [1] iot: Identify the asset name for transformer 'TX1'
       tool: assets  args: {'site_name': 'MAIN'}
       deps=none | expected: List of assets at the site 'MAIN',
       including the id and name of 'TX1'
  [2] fmsr: Assess the winding temperature for transformer 'TX1'
       tool: assess_winding_temperature
       args: {'asset_name': 'TX1', 'wti': 80, 'oti': 85,
              'ati': 78, 'oti_a': 2, 'oti_t': 1}
       deps=#1 | expected: Assessment result for the winding
       temperature of 'TX1'
 
Execution History
  [OK] Step 1 (iot): Identify the asset name for transformer 'TX1'
       tool: assets  args: {'site_name': 'MAIN'}
       {
         "site_name": "MAIN",
         "total_assets": 1,
         "assets": ["Chiller 6"],
         "message": "found 1 assets for site_name MAIN."
       }
  [OK] Step 2 (fmsr): Assess the winding temperature for transformer
       'TX1'
       tool: assess_winding_temperature
       args: {'asset_name': 'TX1', 'wti': 80, 'oti': 85,
              'ati': 78, 'oti_a': 2, 'oti_t': 1}
       {
         "asset_name": "TX1",
         "thermal_status": "Normal",
         "hot_spot_rise_c": 2.0,
         "ageing_rate": 1.039,
         "alarm_active": true,
         "trip_active": false,
         "risk_level": "Low",
         "reasoning": "The wind..."
       }
 
Answer
  The winding temperature for transformer 'TX1' is within the normal
  range with a thermal status of "Normal" and a risk level of "Low".
  The recommended action is to continue monitoring temperature
  readings and perform routine maintenance.
\end{Verbatim}
\end{promptbox}

\begin{promptbox}{Transformer Load profile assessment}
\begin{Verbatim}
$ uv run plan-execute --show-plan --show-history "Assess the load
  profile for transformer with the following readings: Voltages:
  VL1 110, VL2 115, VL3 108, Currents: IL1 100, IL2 105, IL3 95,
  Line-to-line voltages: VL12 225, VL23 220, VL31 218,
  Neutral current: 5, Rated MVA: 100"
 
Plan
  [1] fmsr: Assess the load profile for the given transformer
       readings
       tool: assess_load_profile
       args: {'asset_name': 'transformer', 'vl1': 110, 'vl2': 115,
              'vl3': 108, 'il1': 100, 'il2': 105, 'il3': 95,
              'vl12': 225, 'vl23': 220, 'vl31': 218,
              'inut': 5, 'rated_mva': 100}
       deps=none | expected: Load profile assessment result for the
       transformer
 
Execution History
  [OK] Step 1 (fmsr): Assess the load profile for the given
       transformer readings
       tool: assess_load_profile
       args: {'asset_name': 'transformer', 'vl1': 110, ...}
       {
         "asset_name": "transformer",
         "load_mva": 19.24,
         "load_factor_pct": 19.24,
         "loading_status": "Normal",
         "current_imbalance_pct": 5.26,
         "neutral_current_flag": false,
         "reasoning": "The lo..."
       }
 
Answer
  The transformer's load profile is normal with a load factor of
  19.24% and a current imbalance of 5.26%. The loading status is
  within safe operating parameters, and it is recommended to
  continue monitoring the transformer's condition.
\end{Verbatim}
\end{promptbox}

\end{document}